\documentclass[journal]{IEEEtran}
\usepackage[utf8]{inputenc}

\usepackage{algorithm}
\usepackage{algorithmic}
\usepackage{amsmath}
\usepackage{amssymb}
\usepackage{array}
\usepackage{bm}
\usepackage{booktabs}
\usepackage{cases}
\usepackage[pdftex]{graphicx}
\usepackage{indentfirst}
\usepackage{mathrsfs}
\usepackage{multirow}
\usepackage{url}
\usepackage{xcolor}
\usepackage{makecell}
\usepackage{pifont}
\usepackage{color}  
\usepackage{subfigure}
\usepackage{underscore}
\usepackage{soul}

\usepackage[colorlinks,
            linkcolor=red,
            anchorcolor=black,
            citecolor=green]{hyperref}
\newcommand{\eg}{\textit{e.g.},~}
\newcommand{\ie}{\textit{i.e.},~}

\begin{document}

\title{Temporal Aggregation for Adaptive RGBT Tracking }
%
\author{Zhangyong~Tang,
        Tianyang~Xu,
        and Xiao-Jun~Wu$^*$ 
\thanks{Z. Tang, T. Xu and X.-J. Wu (Corresponding Author) are with the School of Artificial Intelligence and Computer Science, Jiangnan University, Wuxi, P.R. China. (e-mail: \{zhangyong\_tang\_jnu; tianyang\_xu; xiaojun\_wu\_jnu\}@163.com)}

}
%
%
%
%
%
\maketitle
\sloppy
\begin{abstract}
Visual object tracking with RGB and thermal infrared (TIR) spectra available, shorted in RGBT tracking, is a novel and challenging research topic which draws increasing attention nowadays.
In this paper, we propose an RGBT tracker which takes spatio-temporal clues into account for robust appearance model learning, and simultaneously, constructs an adaptive fusion sub-network for cross-modal interactions.
Unlike most existing RGBT trackers that implement object tracking tasks with only spatial information included, temporal information is further considered in this method.
Specifically, different from traditional Siamese trackers, which only obtain one search image during the process of picking up template-search image pairs, an extra search sample adjacent to the original one is selected to predict the temporal transformation, resulting in improved robustness of tracking performance.
As for multi-modal tracking, constrained to the limited RGBT datasets, the adaptive fusion sub-network is appended to our method at the decision level to reflect the complementary characteristics contained in two modalities.
To design a thermal infrared assisted RGB tracker, the outputs of the classification head from the TIR modality are taken into consideration before the residual connection from the RGB modality.
Extensive experimental results on three challenging datasets, \ie VOT-RGBT2019, GTOT and RGBT210, verify the effectiveness of our method.
Code will be shared at \textcolor{blue}{\emph{https://github.com/Zhangyong-Tang/TAAT}}.
\end{abstract}
%
%
%

\section{Introduction}\label{introducion}
With the bounding box of the first frame provided, visual object tracking is committed to predicting the state (location and scale) of a specific object in subsequent frames.
Many industries and military applications, such as surveillance \cite{surveillance} and unmanned aerial vehicle \cite{uav}, attract ascendant attention, boosting visual object tracking a hot topic of pattern recognition and computer vision.
In general, tracking with the RGB modality is the most widely studied sub-task in the tracking community. 
RGB data is supposed to have more perceptual and clear texture information, as it is imaged with the visible lights reflected by the object and background.
So all the factors that affect light reflection exhibit a great impact on the quality of the obtained data.
Extremely, in the night scene, the appearance of an object is almost unseen and naturally leads to tracking failure.
On the other side, thermal infrared sensors are employed to capture the infrared radiation emitted by the object with a wavelength between 3 and 15 microns. 
The quality of TIR data only depends on the object itself, and in other words, TIR data is insensitive to the environmental elements, \eg humidity and illumination condition.
However, TIR data suffers from the thermal crossover.
From the above description, it is apparent that the RGB and TIR modalities complement each other, which is consistently investigated in multi-spectral information fusion \cite{li2020nestfuse, luo2016novel}, and tracking with both modalities is called RGBT tracking.
Commonly, the key point of processing multi-modal data is located on the fusion of multi-modal representations. 
In terms of the fusion stage, there are three fusion mechanisms, \ie pixel-level fusion \cite{pixel-fusion2007}, feature-level fusion \cite{CAT, CMPP, CBPNet} and decision-level fusion \cite{mfDimp}.
First, fusion at pixel level achieves data interaction at the input stage.
The images are  straightforwardly averaged or concatenated in a pixel-wise manner \cite{mfDimp}.
However, when capturing multi-modal data, the cameras are not fixed in the same place.
Thus the manual adjustment plays an essential role in the image alignment procedure.
In other words, it is supposed to have a subtle offset between the paired RGB and TIR images. 
Therefore, the pixel-wise operation would deliver the phenomenon of double image and bring in some extra noise.
Second, with the development of deep learning, feature-level fusion has become the most widely used fusion strategy in the RGBT tracking community.
Compared with pixel-level fusion, the offset can be relieved by the downsampling operation which is bound to show up in all advanced deep neural network architectures.
At this fusion stage, the multi-modal interactions can be roughly separated into two types, \ie feature extraction accompanied with fusion and feature extraction before fusion.
For example, considering the modality-specific challenges, CAT \cite{CAT} achieves cross-modal feature transformation from the first block to the last one.
On the other side, CMPP \cite{CMPP} employs an attention mechanism to fulfil cross-modal fusion given extracted features.
\begin{figure}
	\begin{center}
		\includegraphics[width=1\linewidth, height=0.43\linewidth]{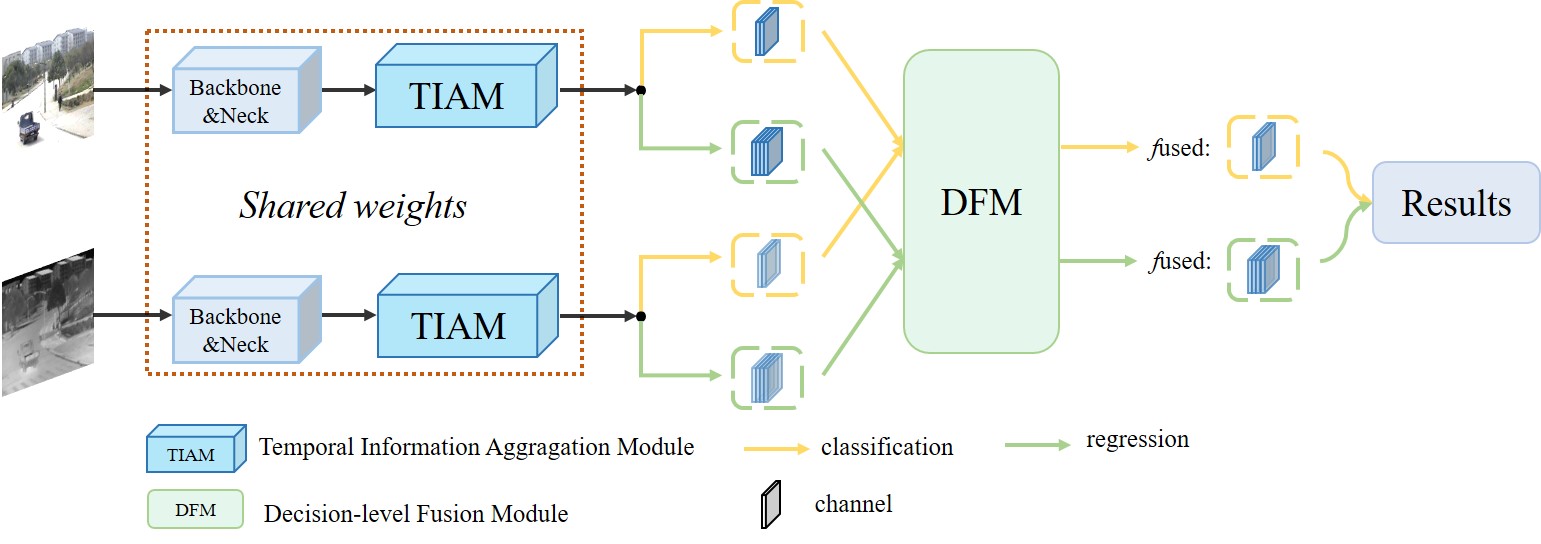} 
	\end{center}
    \caption{Illustration of the proposed multi-modal tracker TAAT. RGB and TIR images are firstly input to the feature extraction blocks (Backbone\&Neck). Then the temporal information is incorporated within TIAM. After the tracking results are obtained within each modality, DFM is followed to achieve adaptive cross-modal fusion at the decision level. 'channel' with light blue represents results from TIR modality while these from RGB modality are tinting by dark blue.}
    \label{fig:pipeline}
\end{figure}
Generally, feature-level fusion is supposed to achieve promising results.
However, the limited RGBT data suppresses the probability of designing a wider and deeper neural network, while lightweight architectures can not extract the multi-modal features with adequate discrimination.
Due to the less discriminative features, fusion at the decision level, which has not been sufficiently studied yet, is another advisable substitution adopted in our method.
Its effectiveness has been proved in the VOT-RGBT2020 challenge \cite{VOT2020} as DFAT \footnote{https://github.com/Zhangyong-Tang/DFAT}, whose cross-modal fusion step is also realized at the decision level, ranks the first.

Compared to the imaged-based tasks \cite{li2017multi, luo2017image, feng2017face, wu2004new, zheng2006nearest, chen2018new, li2011no, wang2003initial, sun2011quantum, sun2019effective}, the sequential relationship is of great importance for video-based topics, such as object tracking and action recognition.
However, existing tracking algorithms pay major attention to learn a robust appearance model spatially and seldomly consider combining the spatial appearance with temporal information.
Although these methods achieve good results in both RGB and RGBT tracking \cite{LADCF,GFSDCF,FANet,DAPNet}, the temporal relationship should not be ignored and is supposed to work positively for more robust trackers.
For instance, the long-term context propagation module in CMPP \cite{CMPP} also verifies the conjecture aforementioned.

Motivated by the above observations, we propose \textbf{\textit{T}}emporal \textbf{\textit{A}}ggregation for \textbf{\textit{A}}daptive RGBT \textbf{\textit{T}}racking (\textbf{\textit{TAAT}}).
Firstly, a lightweight network module (DFM) is designed to achieve cross-modal fusion at the decision level.
Specifically, for classification branches, a fusion matrix is further learned after the corresponding multi-modal results are obtained.
In this way, the latent noise introduced by the biased collaboration can be well relieved.
And a lightweight architecture trained with small scale RGBT data is sufficient for the basic purpose of fusion. 
Compared with the simple average, the influence of distractors is retarded as shown in Fig. \ref{fig:comparison-dfm}, which leads to better tracking performance.
Secondly, during training the Siamese based trackers, one template image is traditionally paired with one search image for each modality for target matching.
In our method, another search image is further selected before the original search image within a range of 5 frames.
Our Temporal Information Aggregation Module (TIAM) is trained to accomplish the feature enhancement task by taking the features of both search images into account.
In this way, the temporal relationship can be explicitly modelled and provides more discriminative feature representations, as shown in Fig. \ref{fig:comparison-tiam}.



Our contributions can be summarized as follows:
\begin{itemize}
	\item Considering the biased collaboration and limited RGBT data, we design an adaptive fusion module at the decision level (DFM) to improve the robustness against distractors.
	\item To integrate temporal information, we design a novel training strategy to train our TIAM and further model the sequential relationship, delivering more discrimination.
	\item The experimental results on three challenging datasets,\ie GTOT \cite{GTOT}, RGBT210 \cite{RGBT210} and VOT-RGBT2019 \cite{VOT2019}, verify the effectiveness of our method. 
\end{itemize}

The rest of this paper is arranged as follows. 
The related works are briefly introduced in Section II. 
The next section, Section III, gives a detailed description of our method. 
And then the multi-step training strategy and the extensive experimental results are exhibited and analysed in Section IV. 
Finally, in Section V, an elaborate statement is drawn to conclude the proposed method.

\section{Related Work}\label{related work}
\subsection{Cross-modal Fusion mechanisms}
As mentioned before, existing cross-modal fusion mechanisms can be divided into three categories, \ie pixel-level fusion, feature-level fusion and decision-level fusion.
In this sub-section, we will give a brief description of these fusion strategies.

Typically, \cite{pixel-fusion2006} and \cite{pixel-fusion2007} are the inchoate trackers that execute pixel-level fusion by averaging and Dual-Tree Complex Wavelet (DW-CWT) \cite{DW-CWT}, respectively.
However, due to the latent deviation introduced in the collaborating procedure, only a few trackers explore the fusion capability at pixel level recently. 
For instance, RGB and TIR images are directly concatenated in mfDiMP \cite{mfDimp} while an image fusion method, MDLatLRR \cite{MDLatLRR}, is introduced to incorporate the multi-modal complementary clues in DFAT \cite{VOT2020}.

However, most RGBT trackers pay attention to the feature-level fusion.
In \cite{GTOT}, which releases an RGBT dataset, \ie GTOT, containing 50 videos and its corresponding benchmark, adaptive weights are solved and optimized by the reconstruction formulation. 
Similarly, \cite{lan2018robust} calculates the fusion weights based on LPBoost \cite{LPBoost} under the max-margin principle.
After the RGBT dataset, RGBT234 \cite{RGBT234}, is published, a series of deep-learning-based algorithms are proposed.
It is worth noting that another dataset, RGBT210 \cite{RGBT210}, without sequences in hot days, is a sub-set of RGBT234.
To reformulate the fusion task, \cite{Chenglong2018Fusing} employs denoising objective to calculate channel-wise reliability by simultaneously considering multi-modal features, discarding less discriminative channels.
Although each channel is also endowed a representative weight, based on MDNet \cite{MDNet}, FANet \cite{FANet} collaborates all channels and designs an adaptive fusion sub-network to achieve cross-modal fusion.
To realize compact cross-modal interactions, DAPNet \cite{DAPNet} starts fusion at the beginning of feature extraction, involving multi-layer features.
Further, instead of the simple convolutional layer used in DAPNet, DAFNet \cite{DAFNet} calculates fusion weights by the softmax operation embedded in an adaptive fusion module.
For more accurate representation, MANet \cite{MANet} constructs generality, modality and instance adapters to respectively catch the modality-shared, modality-specific and instance-specific features, with the multi-modal information being interacted in the generality adapter.
Considering the disgruntling tracking efficiency caused by online updating the fully connected layers, SiamFT \cite{SiamFT} primarily adopts the Siamese network architecture \cite{SiamFC} for RGBT tracking.
The template features from RGB and TIR modalities are concatenated directly while the search features are concatenated after the employment of fusion weights.
Since feature maps from different deep layers contain different semantics, based on the architecture of Dynamic Siamese network \cite{Dynamic-Siam}, DSiamMFT \cite{DSiamMFT} employs the feature maps of each layer as the input of a SiamFT tracker, and aggrandizes a tracker-level fusion to realize the aggregation of multi-resolution clues from different layers. 
Besides, the fusion matrix in CMPP \cite{CMPP} is derived in the form of self-attention for cross-modal pattern propagation.
On the contrary, CAT \cite{CAT} carries out the fusion task as well as feature extraction jointly since the modality-specific features are different from the first block to the last one.
In CBPNet \cite{CBPNet}, channel attention and bilinear pooling \cite{bilinear-pooling} mechanisms are utilized to improve tracking accuracy within each modality.
Then, a quality-aware fusion module is followed to integrate multi-modal superiority.
All the deep-learning-based trackers mentioned above merely preserve the fused features or the features from each modality.
To retain the original features after fusion, \cite{MaCNet} performs multi-modal multi-layer fusion with the weights computed from the original input images.
Similarly, TFNet \cite{TFNet} uses a trident fusion network to better excavate the complementary characteristics of RGB and TIR modalities.
Different from the lightweight network utilized in most RGBT trackers, such as \cite{CMPP, FANet, DAPNet,DAFNet}, mfDiMP \cite{mfDimp} extends DiMP \cite{Dimp} to multi-modal tracking and uses the dataset synthesized from GOT10K \cite{GOT-10K} by \cite{ECO-tir} as its training set.

Fusion at the decision level draws less attention compared with fusion at the feature level.
However, it has shown great potential in RGBT tracking.
By employing the standard addition operation, the decision-level fusion in mfDiMP \cite{mfDimp} is realised efficiently.
In \cite{IPT2019-decision}, KL Divergence is used to fuse the response maps from correlation filter-based (both RGB and TIR modalities considered) and histogram-based (only TIR modality employed) tracking modules.
Analogously, \cite{VCIR2020-decision} incorporates the multi-modal information with the modality reliability calculated by KL Divergence.
Partially focusing on the fusion sub-network, JMMAC \cite{JMMAC} learns the fusion matrix after RGB and TIR image patches go through a fusion network.
JMMAC ranks first in both the published datasets of VOT-RGBT2019 and VOT-RGBT2020 challenges.
What's more, DFAT, the champion of the VOT-RGBT2020 challenge \cite{VOT2020},  also achieves cross-modal fusion  at the decision level by easing the phenomenon caused by data bias.

In this paper, an adaptive decision-level fusion block is designed to accomplish the cross-modal fusion task. 
Different from the hand-crafted features, such as Color Names (CN) \cite{CN} and Histogram of Oriented Gradients (HOG) \cite{HOG}, used in \cite{IPT2019-decision,VCIR2020-decision}, ResNet-50 \cite{Resnet50} is adopted in our method to improve the robustness of feature representations.
Although ResNet-50 is also introduced in mfDiMP, it mainly focuses on feature-level fusion and its fusion strategy at the decision level can be further developed.
DFAT achieves promising performance with its de-biasing fusion block.
However, its hand-crafted fusion scheme just takes the locations with classification scores greater than 0 into account.
After further analysis, it is proved in our method as shown in Fig. \ref{fig:leakyrelu} that negative features can still contribute to the final performance.
The fusion weights in our method are calculated from the classification scores while JMMAC \cite{JMMAC} obtains those from its original input image patches which abets the time consumption.
In summary, we propose an adaptive as well as speedy fusion module equipped with a superior feature extractor.

\subsection{Learning of Object Appearance}
Following the tracking-by-detection paradigm, almost all the existing trackers execute the trajectory prediction mission by constructing appearance models.
Among these algorithms, the construction of the appearance model can be divided into two categories according to its manner.
The first type is learning with only spatial information involved.
A series of MDNet \cite{MDNet} based trackers, \eg \cite{MANet,DAPNet,DAFNet,CBPNet,TFNet,FANet}, belong to this category.
They firstly extract feature representations from the current multi-modal image pair and then use the online updated fully connected layers to explicitly distinguish the target-specific clues.
Similarly, hand-crafted features, \eg HOG and CN, of the current search image are utilized for appearance construction in \cite{Neurocomputing2019,IPT2019-decision,VCIR2020-decision,TNNLS-2021}.
In addition, JMMAC \cite{JMMAC}, which employs ECO \cite{ECO} as its appearance tracker, draws both hand-crafted and depth clues to improve tracking accuracy and robustness.
Besides the spatial-only models, another category is learning with both spatial and temporal information.
To the best of our knowledge, DSiamMFT \cite{DSiamMFT} and CMPP \cite{CMPP} are the only two RGBT trackers that can be concluded into this type.
In DSiamMFT, regularized linear regression is used to learn the feature-level appearance variation transformation between features of the first and previous frames.
When dealing with the current frame, the appearance variation transformation is applied to make the template almost the same as the one extracted from the previous frame.
As mentioned before, in CMPP, the fusion matrix is calculated by the spatial attention mechanism.
After fusion, the fused features of the previous four frames are fetched from the cache and concatenated together.
Similarly, temporal information is integrated as the attention mechanism used in inter-modal fusion. 

In this paper, we propose a Temporal Information Aggregation Module (TIAM).
Compared with DSiamMFT \cite{DSiamMFT}, the template is fixed since the first frame is the only groundtruth given in visual object tracking and of great importance for further guidance.
CMPP \cite{CMPP} uses 5 frames for temporal consistency learning, which costs considerable memory for data storage and time for model training and inferring.
However, not only the accurate bounding box prediction is significant but also the real-time performance values a lot.
So unlike CMPP \cite{CMPP}, only the feature representations of the previous frame are considered in our method.
In this way, the spatio-temporal clues are combined and the feature representations are supposed to be more concentrated on the certain object. 

\section{Proposed Method}\label{proposed method}
In this section, we will firstly give a brief description of the basic tracker SiamBAN \cite{SiamBAN}.
Then, a thorough introduction of our method will be exhibited in the sub-section \ref{tracker}.

\begin{figure}
	\begin{center}
		\includegraphics[width=1\linewidth]{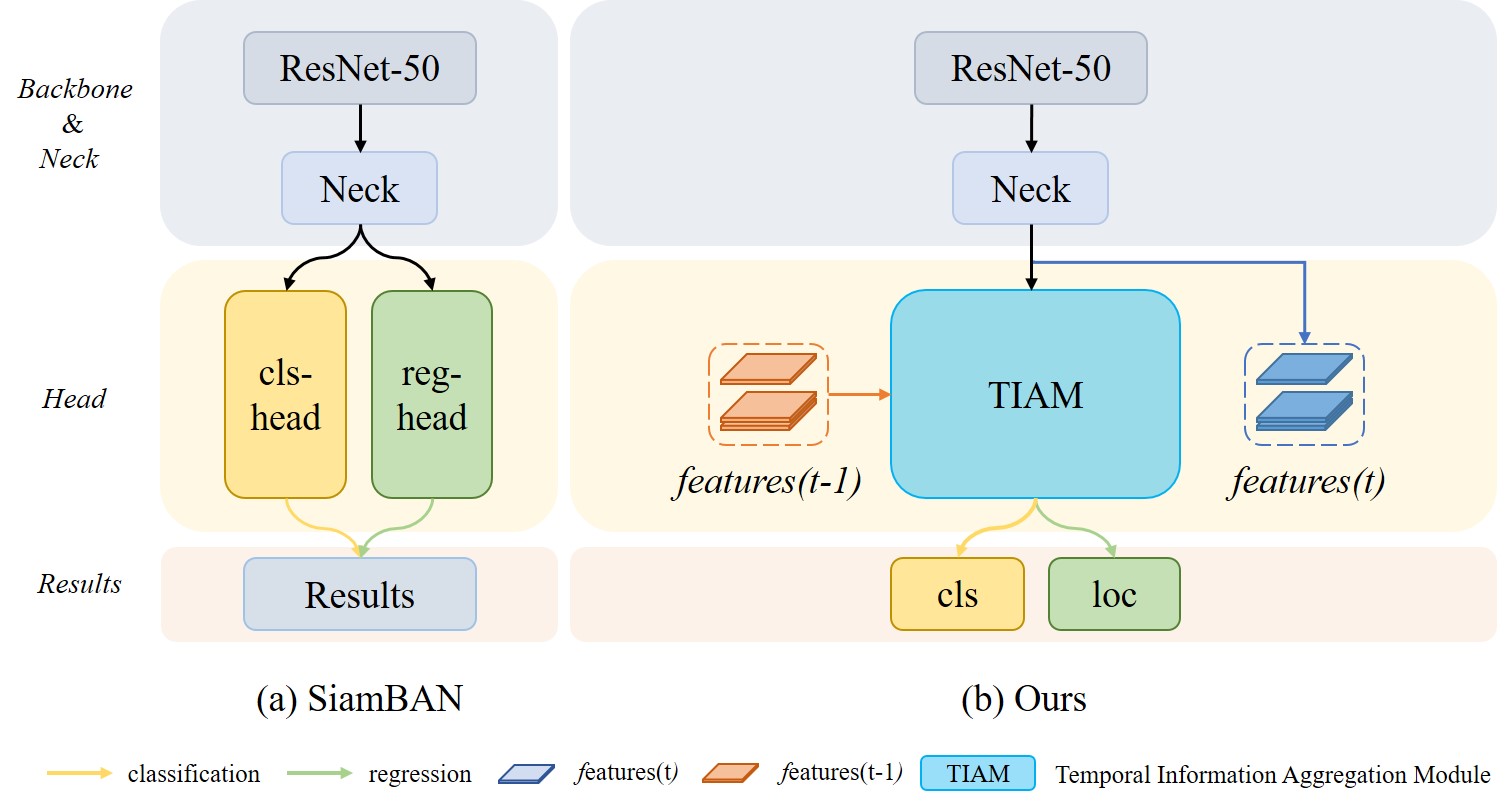} 
	\end{center}
    \caption{Comparison between our baseline tracker (a) and the proposed method (b) in processing the appearance for each single modality}
    \label{fig:ComparisonWithBaseline}
\end{figure}

\subsection{Siamese Tracker}\label{baseline}
The basic purpose of Siamese based trackers is to learn a general representation to measure the similarity between template and search image patches.
In particular, firstly, the image pair is cropped and resized to cater for the input of the network.
After the original image signal passing through the backbone, such as ResNet \cite{Resnet50}, AlexNet \cite{AlexNet} or VGGNet \cite{VGGNet}, 'Neck' block is followed for dimension reduction.
Then the features are fed into the 'Head' block for classification and accurate bounding box estimation.
The outputs of the 'Head' block contains two parts, \ie the results of classification and regression heads.
Finally, after post-processing, which is mainly designed for temporal smoothing, the final tracking result in the form of the rectangle is identified from the results of the regression branch with the maximal classification score.

The architecture of SiamBAN \cite{SiamBAN} is shown in the left side of Fig.\ref{fig:ComparisonWithBaseline}. 
As we can see, ResNet-50 \cite{Resnet50} is employed as the feature extractor.
Originally, similar to SiamRPN++ \cite{SiamRPN++}, three 'Neck' blocks are used to project the features from the second, third and fourth residual layers into a common lower-dimensional space, generating compact feature representations for the following matching task.
After obtaining the results of classification and regression branches, the final bounding box is selected from the regression results according to the index of the location with the highest classification score.
Based on the above structure, to satisfy the real-time requirement, only the features from the third layer and its corresponding 'Neck' block are reserved in our designed approach. 
And it should be noted that, since the template is settled after the first frame, we regard it as the latent clue within the classification head (cls-head) and regression head (reg-head).

The contributions of SiamBAN consist of two parts.
For the classification branch, division of ellipses are used to accurately distinguish the sampled positive, negative and ignored labels, which is of much concern for the discrimination of foreground object and background distractors.
For the regression branch, inspired by FCOS \cite{FCOS}, the traditional fine-tuned hyper-parameters, including aspect ratios and scales of predefined anchors, are abandoned for better generalization.
Besides, the Intersection over Union (IoU) loss is employed to jointly regress the distances from the location to the four sides of the groundtruth bounding box.


\subsection{Temporal Information Aggregated RGBT Tracking}\label{tracker}
In this sub-section, we firstly present the overview network structure.
Then, the details of the proposed modules, \ie Temporal Information Aggregation Module (TIAM) and Decision-level Fusion Module (DFM), are discussed.

\textbf{Network Overview:} As shown in Fig. \ref{fig:pipeline}, features of RGB and TIR images are extracted by the 'Backbone' and 'Neck' modules, which have shared parameters for the RGB and TIR modalities.
After that, with the feature representations of the previous frame integrated, TIAM is followed to achieve feature enhancement with temporal information.
Then the results of the classification and regression branches from both modalities can be obtained.
For cross-modal interactions, all the single-modal results are sent into DFM for the generation of the fused classification and regression response maps.
Consistent with the traditional algorithms, the final tracking result is picked out from the fused regression results where its classification probability ranks the first.

\begin{figure}
	\begin{center}
		\includegraphics[width=1\linewidth, width=1\linewidth]{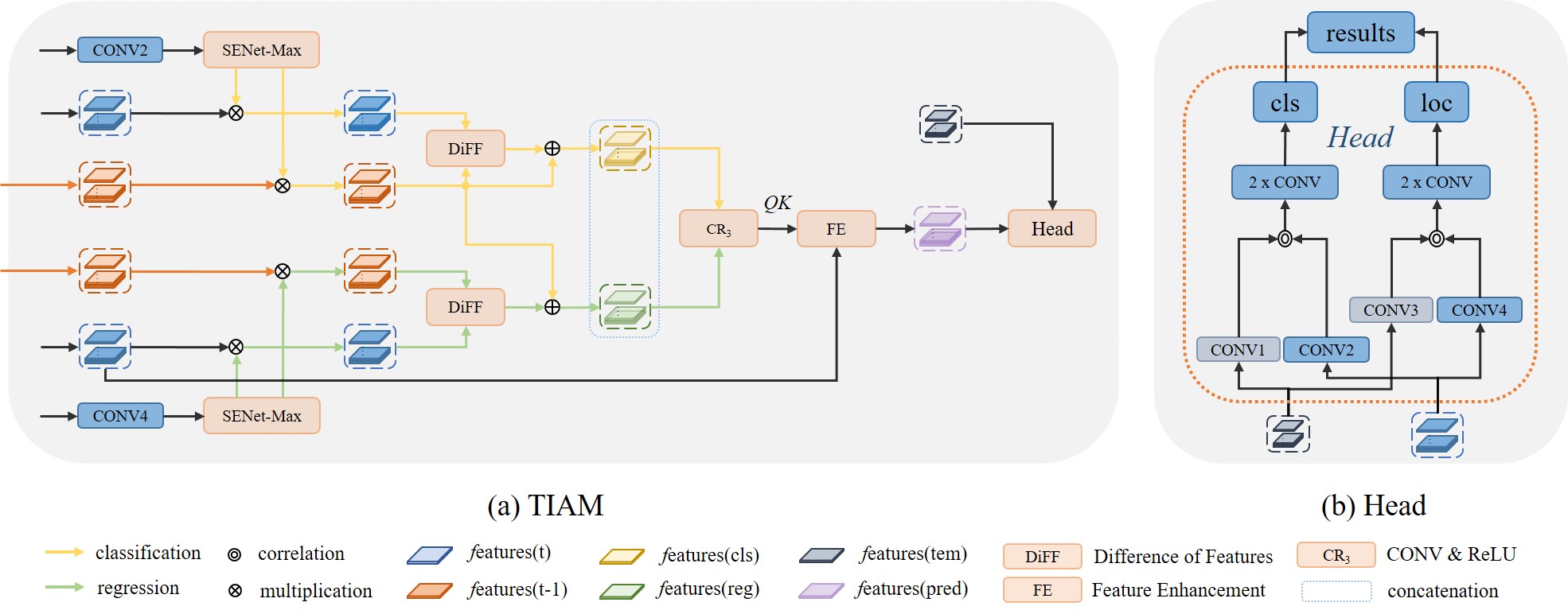} 
	\end{center}
    \caption{Detailed construction of our TIAM (a) and the original Head (b) in SiamBAN.}
    \label{fig:TIAM}
\end{figure}

\textbf{Temporal Information Aggregation Module:}
In principle, exploring temporal information is essential in video analysis.
So the motivation of our Temporal Information Aggregation Module (TIAM) is to endow the temporal interaction in the feature extraction stage to improve the tracking performance.
From Fig. \ref{fig:TIAM}\textcolor{red}{(b)}, it is obvious that the classification and regression heads are independent of each other.
Hence, our TIAM achieves temporal information aggregation with disentangled classification and regression clues.
Fig. \ref{fig:ComparisonWithBaseline} and Fig. \ref{fig:TIAM} give an intuitive comparison between the original prediction heads and our TIAM module.
As for the original heads, similarities between features of the template and the current search patch are calculated after both of them pass the multi-convolutional layers.
Since the existence of potential ambiguity in the calculated similarity matrices, we specially design the first four convolutional layers which are named CONV1, CONV2, CONV3 and CONV4 in Fig. \ref{fig:TIAM}.
The capacity of CONV1 and CONV3 is designed to transfer the original input to a valid space for the assignment of each head, \ie classification and regression.
Respectively, CONV2 and CONV4 are similar to CONV1 and CONV3.
Since TIAM requires establishing sequential relations,  we reuse CONV2 and CONV4 to decouple the features from previous and current frames, as shown in Fig. \ref{fig:TIAM}\textcolor{red}{(a)}.
After the features of the current frame pass CONV2 and CONV4, a variant of SENet \cite{SENet} (SENet-max) is applied to obtain the channel attention vectors.
Specifically, SENet consists of squeeze and excitation operators achieved by Global Average Pooling (GAP) \cite{GAP} and channel-wise multiplication, respectively.
Here Global Max Pooling (GMP) \cite{SENet} is considered to be more appropriate to replace the original GAP since the salient clues are emphasized in the current tracking framework.
In this way, the calculated vectors are employed to disentangle the information for classification and regression from two adjacent frames.

Then, the feature-level differences are computed by DiFF and its formulations are as follows (take classification as an example).
\begin{equation}\label{formulation-DiFF}
\begin{split}
\textbf{\emph{Diff}} & = {\rm DiFF}(\textbf{\emph{cls_t}}, \textbf{\emph{cls_p}})  \\
     & = {\rm CR_2}({\rm CR_1}(\textbf{\emph{cls_t}}) - {\rm CR_1}(\textbf{\emph{cls_p}}))
\end{split}
\end{equation}
The inputs of the DiFF block are the decoupled classification information from both current ($\textbf{\emph{cls_t}}$) and previous frames ($\textbf{\emph{cls_p}}$).
Here '-' denotes the element-wise subtraction.
A basic convolutional block consists of a convolutional layer, a batch normalization layer and a Rectified Linear Unit (ReLU) activation \cite{ReLU}.
However, experimentally, the Rectified Linear Unit activation in our TIAM is further investigated.
So the ReLU activations are highlighted with different subscripts for distinguishment.
Respectively, ${\rm CR_1}$ and ${\rm CR_2}$ represent a convolutional block with the first and second ReLU activations introduced in our TIAM.
For the prediction of the classification clues, its corresponding difference is fused with that from the previous frame.
Given object motion, we argue that combining the regression information from the previous frame directly may degrade the representation power.
So the predicted regression clues are obtained analogously with the classification clues of the last frame.
Then these two synthesised data are concatenated together to generate uniformed feature representations (\textbf{\emph{QK}}) after the application of another convolutional block, \ie ${\rm CR_3}$.
Finally, the feature enhancement (FE) module is designed at the end of the prediction process.
The formulations for FE step are given by:
\begin{equation}\label{formulation-FE}
\begin{split}
\textbf{\emph{Pred}} & = {\rm R_4}({\rm NL}(\textbf{\emph{QK}}, \textbf{\emph{QK}}, \textbf{\emph{f(t)}}))
\end{split}
\end{equation}
$\emph{f}$ means the feature extractor, and $\emph{f(t)}$ is the features of the current frame \emph{t}.
${\rm R_4}$ is the fourth ReLU activation.
In the form of non-local (NL) \cite{non-local} aggregation, the similarity matrix is calculated with the prediction which is treated as the query and key at the same time.
Therefore, the predicted feature representations are supposed to be more robust since the temporal relationship is integrated.
After that, the template and prediction are fed into the original 'Head' block for calculation of the classification and regression results.

\begin{figure}
	\begin{center}
		\includegraphics[width=1\linewidth]{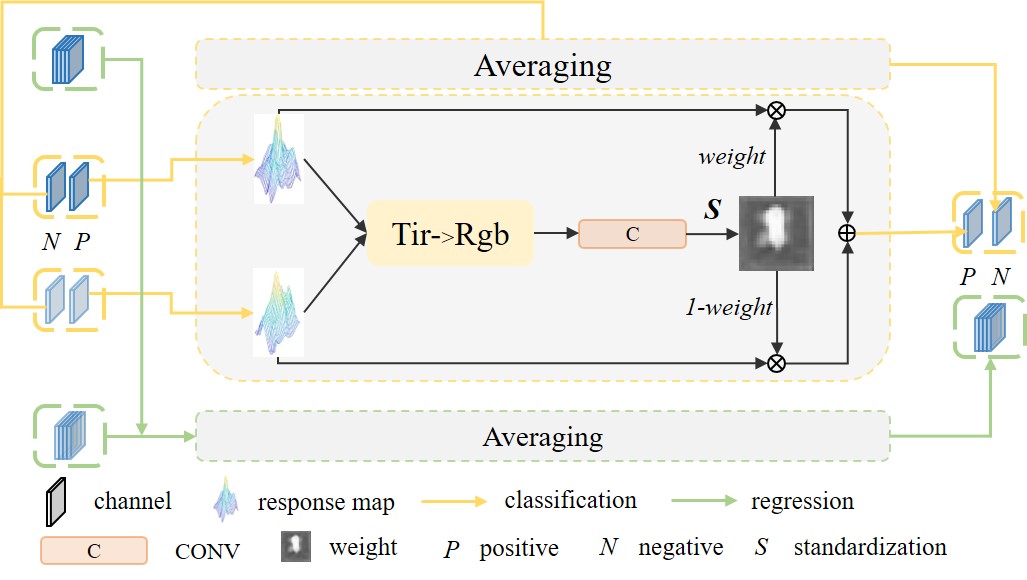} 
	\end{center}
    \caption{Details of DFM. Here 'Tir-\textgreater Rgb' means learning the weight matrix from RGB modality with TIR modality as an assistant.}
    \label{fig:DFM}
\end{figure}

\begin{figure}
	\begin{center}
		\subfigure[Tir-\textgreater Rgb]{
		    \label{tar}
            \includegraphics[width=1\linewidth]{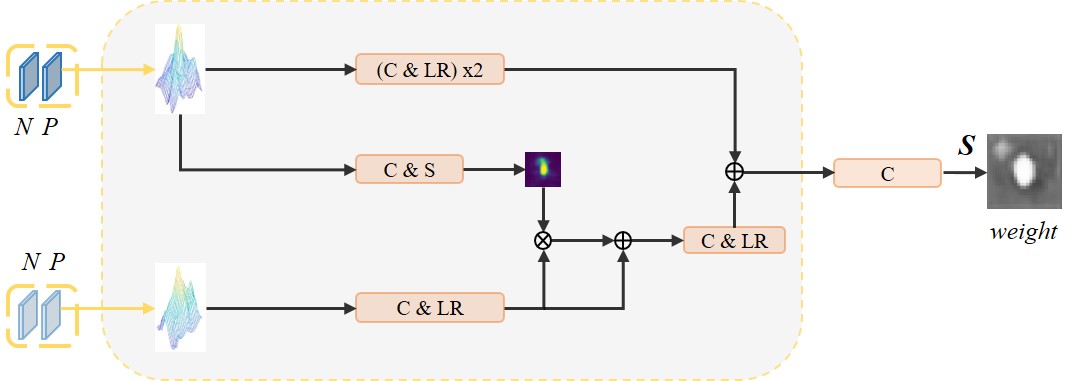}
            }
        \quad
    	\subfigure[Rgb-\textgreater Tir]{
    	    \label{rat}
            \includegraphics[width=1\linewidth]{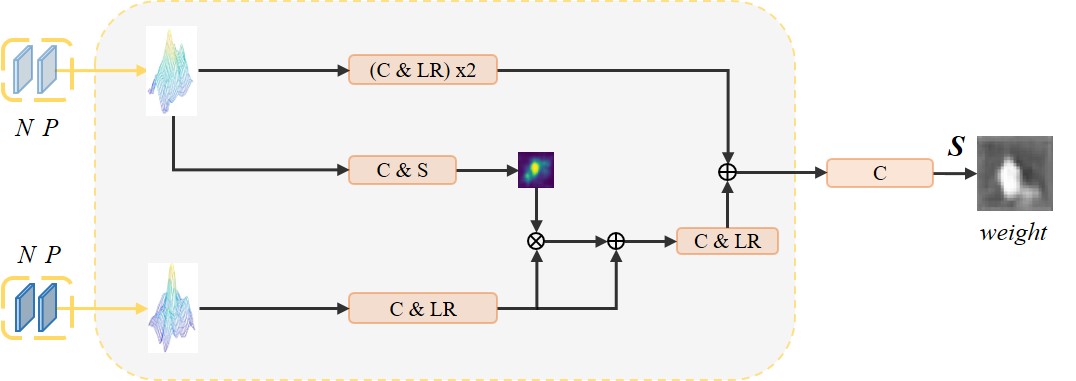}
            }
        \quad
	\end{center}
    \caption{Two variations of the mutual guided block, \ie Tir-\textgreater Rgb (a) and Rgb-\textgreater Tir (b).}
    \label{fig:DFM-detail}
\end{figure}

\textbf{Cross-modal Fusion:}\label{fusion}
As mentioned above, the core of a multi-modal task lies in the combination of multi-modal clues.
In this sub-section, we mainly focus on the novel Decision-level Fusion Module (DFM) proposed in our method.
Most trackers regard object tracking as a binary classification problem, \ie foreground object and background distractors.
The same goes for our algorithm, which means the results of the classification branch are made up of the probabilities of positive and negative samples.
However, based on our observation that the training loss of negative samples, both in the RGB and TIR modalities, stucks in a small amplitude, as shown in TABLE \ref{fig:loss}.
To this end, we ease our DFM to merely focus on fusing the scores of positive samples.
In the centre region of Fig. \ref{fig:DFM}, the response maps of positive sample from both modalities are firstly fed into a sub-block, \ie 'Tir-\textgreater Rgb'.
'Tir-\textgreater Rgb' represents, giving priority to the RGB modality, using the TIR modality to assist RGB modality during the fusion matrix learning process, and vice versa.
Take 'Tir-\textgreater Rgb' for an example, as shown in Fig. \ref{tar}, an attention map is firstly obtained from RGB modality to guide the learning of TIR modality.
Considering that the multiplication will erase the salient characteristics included in each input response map if their peaks locate in different positions, a skip connection is added for the lower branch, which is avoided becoming meaningless, after the multiplication.
The denoised results from the TIR modality are further utilized to guide the learning of the RGB modality.
Then, the fusion matrix is acquired after standardization and sigmoid operators.
The mathematical description is presented as follows.

\begin{equation}\label{formulation-Fusion}
\begin{split}
\textbf{\emph{W_P}} & = {\rm S}({\rm CONV}({\rm Tir->Rgb})) \\
\textbf{\emph{F_cls}} & = \textbf{\emph{R\_cls}} * \textbf{\emph{W_P}} + \textbf{\emph{T\_cls}} * (\textbf{\emph{I}} - \textbf{\emph{W_P}})
\end{split}
\end{equation}
Here CONV means the convolutional layer.
Respectively, \textbf{\emph{R\_cls}} and \textbf{\emph{T\_cls}} represent the classification results from RGB and TIR modalities.
\textbf{\emph{I}} is a matrix with all elements equal to one.
S consists of standardization and sigmoid operators.
Generally, normalization, such as softmax and sigmoid, is considered to constrain the value of weights to [0,1].
However, according to formulation (Eq. \eqref{formulation-Fusion}), the learned fusion matrix $\mathbf{W}$ is only enforced to the RGB modality rather than both RGB and TIR modalities. 
Therefore, during the learning procedure, the matrix normalization should focus on the corresponding single element.
Since softmax operates on balancing multiple elements for vectors, sigmoid is chosen as the normalization operator to adjust the amplitude for each element in the matrix.
Furthermore, a standardization operator is helpful to avoid too many elements locating in both sides of the 'S' shape curve with small gradient. 

\begin{table*}[ht]
\renewcommand\arraystretch{1.2}
\centering
\caption{\label{res:vot19}Quantitative comparison on VOT-RGBT2019 dataset. The top three results of each metric are highlighted in \textcolor{red}{red}, \textcolor{blue}{ blue} and \textcolor{yellow}{yellow}.}
\begin{tabular}{c|cccccccc|c}
\toprule
\toprule
Method& FANet\cite{FANet} & TFNet\cite{TFNet} & MANet\cite{MANet} & mfDiMP\cite{mfDimp} & SiamDW\_T\cite{SiamDW} & ADRNet\cite{ADRNet} & DFAT\cite{VOT2020} & MFNet\cite{JMMAC} & TAAT \\
\toprule
Published& TIV2021 & TCSVT2021 & ICCVW2019 & ICCVW2019 & CVPR2019 & IJCV2021 & - & TIP2021 & -  \\
\midrule
A ($\uparrow$)& 0.4724 & 0.4617 & 0.5823 & 0.6019 & 0.6158 & 0.6218 & {\color[HTML]{FE0000} 0.6652} & {\color[HTML]{3166FF} 0.6465} & {\color[HTML]{FFC702} 0.6426} \\
\midrule
R ($\downarrow$) & 0.4922 & 0.4064 & 0.2990 & {\color[HTML]{FE0000} 0.1964} & {\color[HTML]{3166FF} 0.2161} & 0.2433 & 0.2539 & {\color[HTML]{FFC702} 0.2165} & 0.2647 \\
\midrule
EAO ($\uparrow$) & 0.2465 & 0.2878 & 0.3463 & 0.3879 & 0.3925 & 0.3956 & {\color[HTML]{FFC702} 0.3986} & {\color[HTML]{3166FF} 0.4116} & {\color[HTML]{FE0000} 0.4164} \\
\bottomrule
\bottomrule
\end{tabular}
\end{table*}

\section{Experiments}\label{experiments}
\subsection{Platform Setup}
Our method TAAT is implemented with Pytorch 0.4.1 on a platform with Intel Core i9-9980XE CPU and NVIDIA GeForce RTX 2080Ti GPU. 
Under this situation, our method TAAT achieves real-time performance and reaches 34 frames per second (fps) on the VOT-RGBT2019 benchmark. 

\subsection{Training Strategies}\label{sec:trainingstrategy}
ResNet-50 \cite{Resnet50} is equipped as the backbone in our method.
In general, our training process is separated into three steps, \ie the training of baseline tracker, TIAM and DFM. 

Based on the RGB tracker SiamBAN \cite{SiamBAN}, our RGBT tracker is formed by introducing multi-modal complementarities.
In the first stage, we mainly train the 'Neck' block, classification and regression heads.
Since the backbone is pre-trained on ImageNet \cite{ILSVRC15}, it is fine-tuned in this process for generating more discriminative features for visual object tracking.
As for the selection of training datasets, we choose the datasets which are commonly used in the tracking community, \ie ImageNet DET \cite{ILSVRC15}, COCO \cite{COCO}, GOT-10K \cite{GOT-10K}, ImageNet VID \cite{ILSVRC15} and LaSOT \cite{lasot}.
Besides, all the parameters are updated with the Stochastic Gradient Descent (SGD) optimizer.
The learning rate is warmed up from 0.001 to 0.005 in the first 5 epochs and is gradually degraded to 0.00005 in the left 15 epochs.
The weight decay and momentum, respectively, are set to 0.0001 and 0.9 to prevent our model from over-fitting.
The batch size is 28 and 1 million image pairs are used for training per epoch.
Following the baseline tracker, our loss function includes two parts, \ie classification loss and regression loss.
As mentioned above, distinguishing the object from the background is deemed to be a binary classification problem and its corresponding loss is cross-entropy with all negative and positive samples considered.
For the regression loss, with only positive samples concerned, different from the weighted L1 loss used in \cite{SiamRPN++}, IOU loss is chosen to train the four offsets between our predictions and the corresponding groundtruth in the meantime.
For more details on the training losses, please refer to \cite{SiamBAN}.

Since temporal information is extracted from consecutive frames, GOT-10K \cite{GOT-10K} and LaSOT \cite{lasot}, which are frame-wise annotated video datasets, are utilized to learn the sequential relationship.
Specifically, different from the traditional training scheme, an extra search image, whose index is within the first five frames before the original one, is selected.
In this way, the temporal information can be learned from this pair and results in more robust feature representations.
To guarantee the effectiveness of our baseline method, at this stage, the parameters of the backbone, 'Neck' and 'Head' are settled.
And we set almost all the hyper-parameters the same with the training of baseline method except the number of image pairs used in one epoch (0.4 million for the training of TIAM).

The last step is to train an adaptive decision-level fusion module (DFM) to achieve the cross-modal fusion task.
Thus, the aforementioned datasets with only RGB data contained are improper for the training at this stage, and GTOT \cite{GTOT} and RGBT210 \cite{RGBT210} are wielded for the training of DFM.
It should be noted that, to avoid damaging the effect of the previously trained model, only the parameters of our DFM are set learnable.
Constraint to the limited RGBT data, the number of image pairs is 0.2 million for each epoch.
Other hyper-parameters are the same as the first training step.
Especially, the LeakyReLU is selected as the activation function and the $\lambda$ is set to 0.1, which is further analysed in sub-section \ref{sec:leakyrelu}.

\subsection{Online Testing}\label{sec:test}
Given the initial frame and the current frame, prediction is obtained by the proposed tracking network via forward passing the backbone, neck, TIAM, and DFM modules.
Similar to existing RGBT tracking settings, both RGB and TIR images are input to our model frame by frame in the online tracking phase.
Differently, due to the temporal consistency being delicately designed by our TIAM, the features of the previous frame are stored and participate in the tracking process of the current frame.

\subsection{Datasets and Evaluation Metrics}\label{sec:datasets}
Our testing datasets include VOT-RGBT2019 \cite{VOT2019}, GTOT \cite{GTOT} and RGBT210 \cite{RGBT210}.

VOT-RGBT2019, provided by the VOT community, containing 60 sequence pairs, is the first RGBT dataset used for competition.
Since RGBT234 \cite{RGBT234} provides separate bounding box annotation for each modality, the label of RGB modality is discarded and only the label of TIR modality is employed as groundtruth, which is inherited by VOT-RGBT2019.
There are three evaluation metrics, \ie Accuracy (A), Robustness (R) and Excepted Average Overlap (EAO).
Accuracy measures the overlap between the estimated bounding box and groundtruth.
Robustness represents the extent of tracking failures.
EAO is a comprehensive indicator simultaneously balancing accuracy and robustness.

GTOT consists of 100 sequences in total, \ie 50 in RGB modality and 50 in TIR modality.
Due to the different imaging principles, the collaborated bounding boxes are inconsistent in these two modalities, which means there exists one groundtruth file for each modality.
Precision and success rate are the evaluation metrics for GTOT.
Specifically, precision measures the distance between the centre of the predicted bounding box and its corresponding label.
The distance below a certain threshold, 5 in the benchmark, indicates an accurate prediction.
Precision rate equals the number of accurate predictions divided by the total number of frames.
Similarly, success rate equals the number of frames with IoU exceeding 0.6, for example, above the total number of the whole dataset.

RGBT210 is a larger RGBT dataset with 210 sequences for each modality.
And its benchmark uses the same evaluation metrics as GTOT as well as the identical thresholds.

\subsection{Quantitative Results}
In this sub-section, we present an intuitive comparison between our method and other RGBT trackers on several benchmarks, including VOT-RGBT2019 \cite{VOT2019}, GTOT \cite{GTOT} and RGBT210 \cite{RGBT210}.

\begin{table}[ht]
\centering
\caption{\label{res:gtot-rgbt210}Quantitative comparison on GTOT and RGBT210 datasets with other Siamese based RGBT trackers. The best results of each metric are highlighted in \textcolor{red}{red}.}
\begin{tabular}{c|cccc}
\toprule
\toprule
Dataset & \multicolumn{2}{c}{GTOT} & \multicolumn{2}{c}{RGBT210} \\
\toprule
Tracker & Precision & Success & Precision & Success \\
\midrule
SiamFT  & 0.8220 & {\color[HTML]{FE0000} 0.7000} & - & - \\
\midrule
DSiamMFT & 0.8200 & 0.6720 & 0.6420 & 0.4360 \\
\midrule
TAAT & {\color[HTML]{FE0000} 0.8580} & 0.6960 & {\color[HTML]{FE0000} 0.7100} & {\color[HTML]{FE0000} 0.4860} \\
\bottomrule
\bottomrule
\end{tabular}
\end{table}

\textbf{On VOT-RGBT2019:}
We compare our method with a series of trackers, including FANet \cite{FANet}, TFNet \cite{TFNet}, MANet \cite{MANet}, mfDiMP \cite{mfDimp}, SiamDW\_T \cite{SiamDW}, DFAT \cite{VOT2020}, MFNet \cite{JMMAC}, and ADRNet \cite{ADRNet}.
The results are shown in TABLE \textcolor{red}{\ref{res:vot19}}.
As we can see, our TAAT achieves promising results on VOT-RGBT2019 among these participators.
Compared with MFNet, which is the second place tracker, our method has an improvement of 0.48\% on EAO.
In addition, DFAT, the champion of the VOT-RGBT2020 challenge, obtains the best tracking accuracy due to its joint predictions of bounding box and segmentation.
However, the hand-crafted fusion strategy designed in DFAT results in the 1.78\% behind on EAO compared to our adaptive fusion network.
Therefore, the overall effectiveness of our method has been verified.


\textbf{On GTOT and RGBT210:}
In these two datasets, SiamFT \cite{SiamFT} and DSiamMFT \cite{DSiamMFT} are regarded as the main competitors since they follow the same Siamese framework as our method.
TABLE \ref{res:gtot-rgbt210} reports the comparative results.
On GTOT, our method, TAAT, surpasses DSiamMFT by 3.8\% and 2.4\% on precision and success rate respectively.
Although SiamFT has a slight 0.4\% lead on success rate, its predictions are less accurate and fall behind our method by about 3.6\%.
On RGBT210, our method exceeds DSiamMFT on both precision and success rate metrics by 6.8\% and 5.0\%, respectively.

\subsection{Self-analysis on TIAM}
Generally, our TIAM module aims to introduce temporal clues with classification and regression features disentangled by channel attention.
In this part, several methods for the calculation of channel attention is investigated.
Besides, four ReLU activations are highlighted with subscripts and  they are experimentally compared.

\textbf{Attention Mechanism:}
Due to the independence of the original classification and regression branches, which is discussed in sub-section \ref{tracker}, we separately integrate the temporal information for the two heads.
As shown in Fig. \ref{fig:TIAM}, channel attention is employed to decouple the classification and regression information from the features after passing the backbone and 'Neck' module.
Considering the generalization of SENet \cite{SENet}, which consists of two components, \ie squeeze and excitation, it is employed in our method.
However, following the entire tracking framework, GAP is replaced by GMP since the position with the most salient classification score is picked out as the newly predicted object centre.
The experimental results in TABLE \ref{res:tiam-attention} also confirm the effectiveness of this modification, named SENet-MAX.
Since both ECANet and FcaNet are modifications of SENet, they are pulled into the comparison of this sub-section.
Specifically, ECANet \cite{ECANet} improves SENet by introducing a 1D convolutional kernel for local communication and keeps the dimension stable during excitation.
FcaNet \cite{FcaNet} is another modified method of SENet by completing feature squeeze in the frequency domain.
From TABLE \ref{res:tiam-attention}, SENet with GMP (SENet-MAX) , which respectively obtains improvements of \textit{0.44\% }and \textit{0.17\%} on EAO and accuracy, performs the best.
The manifestation of ECANet is slightly better than SENet on EAO (\textit{+0.02\%}) since the hyper-parameters, \ie $gamma$ and $b$, employed in \cite{ECANet} are remained in our task while FcaNet shows better superiority compared to ECANet on EAO (\textit{+0.32\%}).
In summary, with the feature aggregation replaced by GMP, SENet shows more potential for visual object tracking as experimentally proved in our method.

\textbf{Non-linear Activation:} 
Except for the DFM, ReLU activation is applied to the rest of our method.
Among all the activations in our TIAM, four of them are emphasised with subscripts, \ie $R_1$, $R_2$, $R_3$ and $R_4$, respectively and appear in Eq. \ref{formulation-DiFF}, Fig. \ref{fig:TIAM} and Eg. \ref{formulation-FE}.
As illustrated in Fig. \ref{fig:TIAM}, the prediction of classification is obtained by combining the relevant features from the current and previous frames.
Specifically, the outputs of 'DiFF' block and the corresponding features from the previous frame are mixed by element-wise addition. 
However, the discrepancy is calculated by element-wise subtraction, which means the value of salient clues of the previous frame is supposed to be less than 0 in the discrepancy.
Thus, without $R_2$, the salient representations from the features of the previous frame are supposed to neutralize beyond expectation after the element-wise addition.
Therefore, in TABLE \ref{res:tiam-relu}, the second ReLU activation $R_2$ is enabled among all the quantitative experiments.

\begin{table}[ht]
\centering
\caption{\label{res:tiam-attention}Comparison of different methods for the learning of channel attention. The best results are emphasised in \textcolor{red}{red}.}
\begin{tabular}{c|ccc}
\toprule
\toprule
Attention Mechanism & A {($\uparrow$)} & R {($\downarrow$)}  & EAO {($\uparrow$)} \\
\midrule
SENet  & 0.6524   & 0.2890   & 0.3941 \\
\midrule
ECANet  & 0.6404   & 0.2965  & 0.3943  \\
\midrule
FcaNet  & 0.6473   & {\color[HTML]{FE0000}  0.2688} & 0.3973  \\
\midrule
SENet-MAX  & {\color[HTML]{FE0000} 0.6541} & 0.2802  & {\color[HTML]{FE0000} 0.3985} \\
\bottomrule
\bottomrule
\end{tabular}
\end{table}

With these four ReLU activations enabled, the best results on all three metrics are obtained which indicates that there still exists noise after features passing the 'Neck' block.

\begin{figure}
	\begin{center}
		\includegraphics[width=1\linewidth,height=0.5\linewidth]{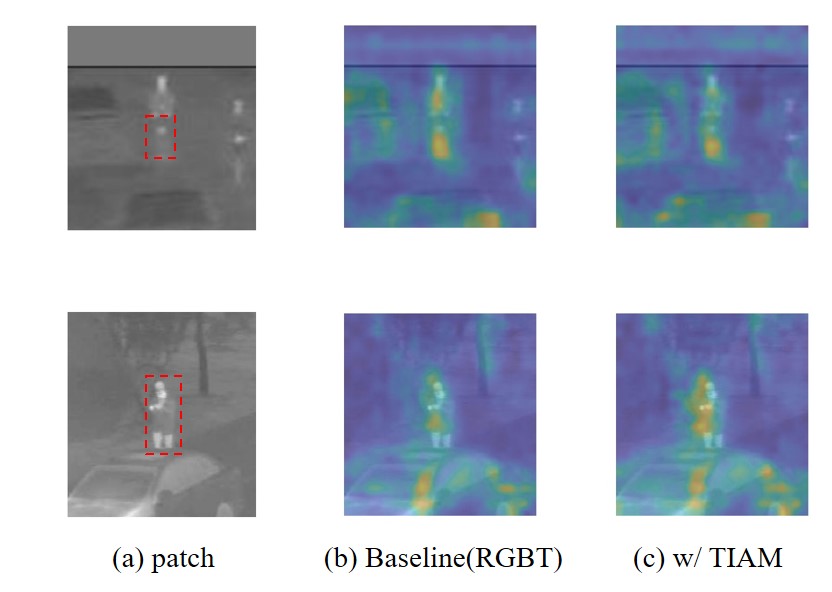} 
	\end{center}
    \caption{Visualisation comparison between our TIAM and baseline tracker. The target bounding boxes are highlighted in red dashed line. The first row is from sequence \textit{baby} and the second is from sequence \textit{man88}, which are from VOT-RGBT2019 dataset.}
    \label{fig:comparison-tiam}
\end{figure}

The qualitative results of our TIAM are displayed in Fig. \ref{fig:comparison-tiam}.
We select frames from sequence \textit{baby} (the first row) and \textit{man88} (the second row) for visualisation. 
The first column shows the original input image patches.
Intuitively, the right two columns exhibit the discrepancy between the baseline and the variant with our TIAM module enabled (w/ TIAM).
For the frame selected from sequence \textit{baby}, our TIAM significantly weakens the nearby surrounding distractors, delivering more discrimination.
Similarly, for the second row, the attention corresponding to the target (man) is also enhanced.
Thus, thanks to the temporal clues integrated by our TIAM, it can be concluded that the learned appearance model supports enhanced discrimination against distractors compared to that constructed with solely spatial appearance.

\subsection{Self-analysis on DFM}
In this part, a detailed experimental analysis of our DFM module is reported.

\begin{figure}
	\begin{center}
		\includegraphics[width=1\linewidth,height=0.5\linewidth]{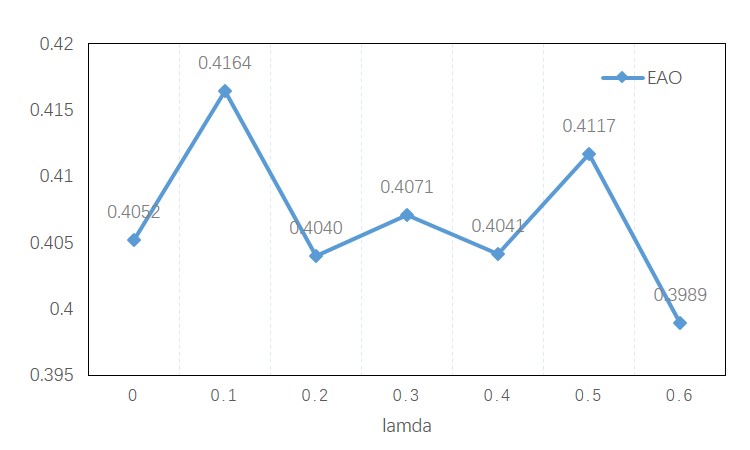} 
	\end{center}
    \caption{Analysis of the LeakyReLU used in DFM.}
    \label{fig:leakyrelu}
\end{figure}

\textbf{LeakyReLU:} \label{sec:leakyrelu}
As mentioned above, different from other modules, LeakyReLU is adopted as the non-linear activation in DFM.
In sub-section \ref{baseline}, it has been known that the final reference for classification is obtained after the normalization between the response maps of positive and negative samples.
In other words, the scores of the same location in both response maps jointly determine the final prediction rather than the numerical values in a single response map.
The inputs of DFM are the responses of positive samples from RGB and TIR modalities.
Generally, the scores for the positive locations can reflect the tracking reliability, which should be above zero.
However, in some challenging situations, such as fast motion and large illumination change, the scores are always lower than usual.
In these cases, the scores below zero can also present useful information since they could be even larger than the above zero ones after normalization.
Totally, the classification results are dominated by positive and sometimes influenced by negative values in the response maps.
Therefore, in our DFM, we use LeakyReLU to maintain the information conveyed by the negative values.

As shown in Fig. \ref{fig:leakyrelu}, the best results are achieved with $lambda$, which is the inherent parameter of LeakyReLU, set to 0.1.
Compared with ReLU ($lambda$ equals 0), an improvement of 2.47\% is obtained.
The performance degrades with the increase of $lambda$.
Generally, it can be seen that a proper influence of negative values enables boosting the tracking performance in challenging clips while maintaining the original robustness to usual scenarios.

\begin{figure}
	\begin{center}
		\includegraphics[width=1\linewidth,height=0.6\linewidth]{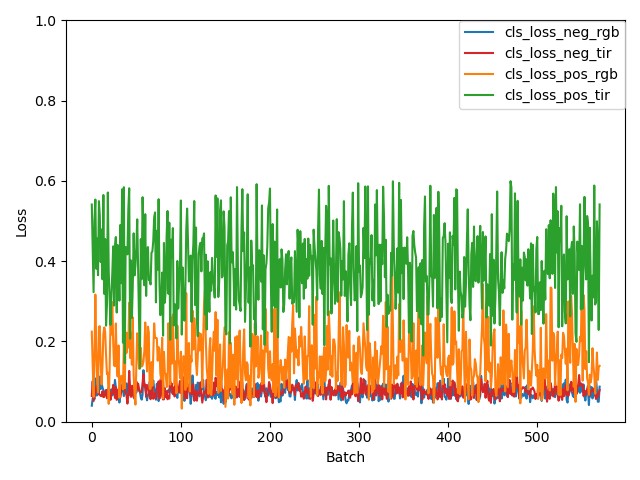} 
	\end{center}
    \caption{Visualization of classification losses from both RGB and TIR modalities for positive and negative samples.}
    \label{fig:loss}
\end{figure}

\textbf{Modality Significance:}
In Fig. \ref{fig:DFM-detail}\textcolor{red}{(a)}, the skip connection for the RGB modality locates in the end.
Thus, this variation is dominated by RGB and supplemented by TIR modality, and vice versa (Fig. \ref{fig:DFM-detail}\textcolor{red}{(b)}).
To further explore the relationship between RGB and TIR modalities, in contrast to our DFM, the other two variations are participated in.
The results are displayed in TABLE \ref{res:dfm-modality}.
Our DFM achieves the best results on both EAO and robustness (R) metrics.
By introducing another variant, which gives priority to the TIR modality, our performance on EAO degrades by 1.01\%.
Furthermore, merely considering the TIR modality as the protagonist, a slight decrease of 0.05\% on EAO appears, which is consistent with our observation based on Fig. \ref{fig:loss} that the discrimination of TIR modality for positive samples is lower than the RGB modality.

\begin{table}[ht]
\centering
\caption{\label{res:tiam-relu}Ablation study for the four ReLU activations on the VOT-RGBT2019 dataset. The best results are emphasised in \textcolor{red}{red}.}
\begin{tabular}{cccc|ccc}
\toprule
\toprule
{\rm $R_1$} & {\rm $R_2$} & {\rm $R_3$} & {\rm $R_4$} & A {$(\uparrow)$} & R {$(\downarrow)$} & EAO {$(\uparrow)$} \\
\toprule
 & \checkmark & & & 0.6488 & {\color[HTML]{3166FF} 0.2806} & {\color[HTML]{FFC702} 0.3893}  \\
\midrule
\checkmark & \checkmark & & & 0.6521 & 0.3046 & 0.3846  \\
\midrule
 & \checkmark & \checkmark & & {\color[HTML]{3166FF} 0.6540} & 0.3172 & 0.3827  \\
\midrule
 & \checkmark & & \checkmark & 0.6517 & 0.3060 & 0.3729  \\
\midrule
\checkmark & \checkmark & \checkmark & & {\color[HTML]{FFC702} 0.6523} & {\color[HTML]{FFC702} 0.2814} & {\color[HTML]{3166FF} 0.3912}  \\
\midrule
\checkmark & \checkmark & & \checkmark & 0.6519 & 0.2940 & 0.3785  \\
\midrule
 & \checkmark & \checkmark & \checkmark & 0.6509 & 0.3084 & 0.3785  \\
\midrule
\checkmark & \checkmark & \checkmark & \checkmark & {\color[HTML]{FE0000} 0.6541} & {\color[HTML]{FE0000} 0.2802} & {\color[HTML]{FE0000} 0.3985} \\
\bottomrule
\bottomrule
\end{tabular}
\end{table}

\begin{table}[ht]
\centering
\caption{\label{res:dfm-modality}Exploring different variations of DFM. The best results are highlighted in \textcolor{red}{red}.}
\begin{tabular}{c|ccc}
\toprule
\toprule
Variation & A ($\uparrow$) & R ($\downarrow$) & EAO ($\uparrow$) \\
\toprule
Tir-\textgreater{}Rgb \& Rgb-\textgreater{}Tir & 0.6408 & 0.2938& 0.4063 \\
\midrule
Rgb-\textgreater{}Tir & {\color[HTML]{FE0000} 0.6475} & 0.2767& 0.4058\\
\midrule
Tir-\textgreater{}Rgb(DFM) & 0.6426 & {\color[HTML]{FE0000} 0.2647} & {\color[HTML]{FE0000} 0.4164}  \\
\bottomrule
\bottomrule
\end{tabular}
\end{table}

\begin{table*}[ht]
\centering
\renewcommand\arraystretch{1.2}
\caption{\label{res:ablation}Ablation studies on VOT-RGBT2019 and GTOT datasets. The detailed introduction of each metric can be found in sub-section \ref{sec:datasets}. Compared with 'Baseline (RGBT)', the improvement of our TIAM, DFM and TAAT are counted on EAO and 'Success' respectively. The top three results are emphasised in \textcolor{red}{red}, \textcolor{blue}{blue} and \textcolor{yellow}{yellow}.}
\begin{tabular}{c|ccc|cc}
\toprule
\toprule
Dataset  & \multicolumn{3}{c|}{VOT-RGBT2019} & \multicolumn{2}{c}{GTOT}\\
\toprule
Evalution Metric & A ($\uparrow$) & R ($\downarrow$) & EAO ($\uparrow$) & Precision ($\uparrow$) & Success ($\uparrow$)\\
\midrule
Baseline(TIR) & 0.6225 & 0.5584 & 0.2439 & 0.7170 & 0.6010\\
\midrule
Baseline(RGB) & 0.5991 & 0.4533 & 0.3197 & 0.7640 & 0.6280\\
\midrule
Baseilne(RGBT) & {\color[HTML]{3166FF} 0.6539} & 0.3505 & 0.3701 & 0.8340 & 0.6780\\
\midrule
w/ DFM & {\color[HTML]{FFC702} 0.6529} & {\color[HTML]{FFC702} 0.3360} & {\color[HTML]{FFC702} 0.3823} (+1.22\%)& {\color[HTML]{FFC702} 0.8460} & {\color[HTML]{3166FF} 0.6880} (+1.0\%)\\
\midrule
w/ TIAM & {\color[HTML]{FE0000} 0.6541} & {\color[HTML]{3166FF} 0.2802} & {\color[HTML]{3166FF} 0.3985} (+2.84\%) & {\color[HTML]{3166FF} 0.8490} & {\color[HTML]{FFC702} 0.6850} (+0.7\%)\\
\midrule
TAAT & 0.6426 & {\color[HTML]{FE0000} 0.2647} & {\color[HTML]{FE0000} 0.4164} (+4.63\%) & {\color[HTML]{FE0000} 0.8580} & {\color[HTML]{FE0000} 0.6960} (+1.8\%)\\
\bottomrule
\bottomrule
\end{tabular}
\end{table*}

\textbf{Network Construction:}
In order to better explore the potential of our DFM, networks with the same depth but different widths have been experimented.
TABLE \ref{res:dfm-network} shows the quantitative results.
Taking experiment '2' as an example, the inputs for DFM are the classification scores of positive samples from each modality and they are both matrices with channel numbers equaling 1.
The dimension of features from the TIR modality after the first convolutional layer is 32 and shrinks to 16 after the second one (lower branch).
Before the residual connection for features from the RGB modality, the response map is enlarged to 16.
After convolution and standardization, the final fusion matrix is obtained with channels reduced to 1.
In order to keep the same receptive field for both modalities before the final addition, an extra convolutional layer for transition is introduced for the RGB modality (upper branch), whose construction is illustrated in Fig. \ref{fig:DFM-detail}\textcolor{red}{(a)}.
Hence, the configuration of '6' means the response map from the RGB modality is firstly broadened to 8 by the transitional layer, and further increased to 16 to match the scale of that from the TIR modality.
From TABLE \ref{res:dfm-network}, it can be seen that wider networks can not deliver better performance since the inputs of DFM are already highly semantic.
As experimentally proved, wider networks ('7', '8') bring more extra noise, while the variation with narrow construction ('1', '2') can not fully excavate the information conveyed from inputs.

\begin{figure}
	\begin{center}
		\includegraphics[width=1\linewidth,height=0.5\linewidth]{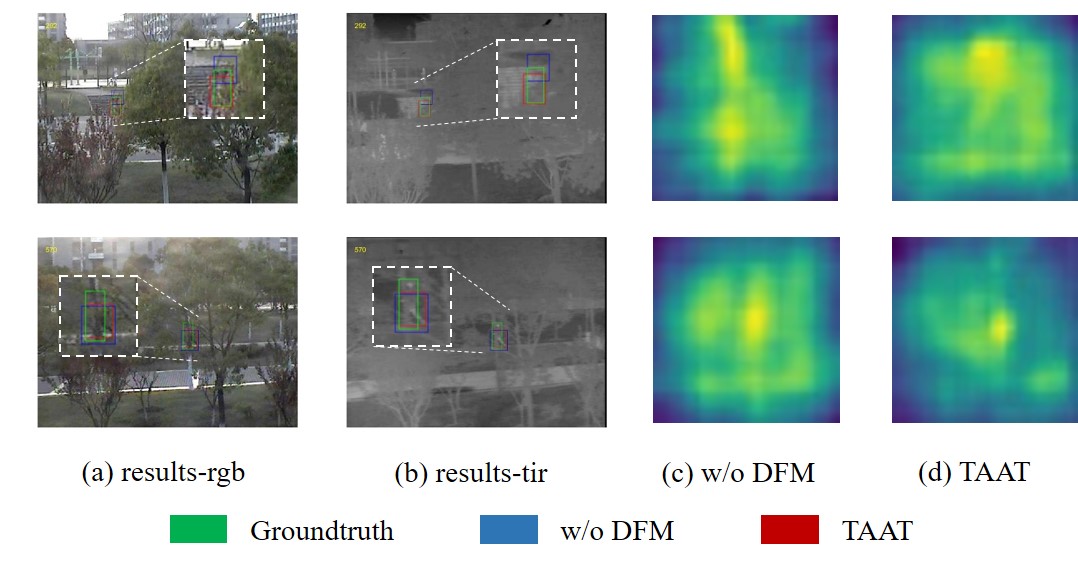} 
	\end{center}
    \caption{Qualitative performance of the proposed 'DFM'. Bounding boxes in green, blue and red correspond to groundtruth, 'w/o DFM' and the proposed TAAT, respectively.}
    \label{fig:comparison-dfm}
\end{figure}

Intuitively, the qualitative comparison of TAAT and 'w/o DFM', in which the multi-modal responses are averaged, is figured in Fig. \ref{fig:comparison-dfm}.
The results of two frames from sequence \textit{car37} in VOT-RGBT2019 \cite{VOT2019} are exhibited for visualization.
The images in the first row are from frame \textit{292} while the second row is from frame \textit{570}.
The first and second column shows the images from the RGB and TIR modalities, respectively, with bounding boxes annotated. 
The third column is the response map for positive samples from the tracker without DFM, 'w/o DFM'.
Similarly, the last column is from our method equipped with DFM.
The target in frame \textit{292} is occluded by leaves with background clutters in both modalities, which means the response within each modality is prone to exhibit multiple peaks. 
After appending our DFM, the multi-modal complementary clues are better fused and the influence of distractors is therefore alleviated.
In addition, the target in frame \textit{570} is partially occluded by trunks and leaves, resulting in a bad appearance in the RGB modality.
Fortunately, since the temperatures of trunks, leaves and the object are different, its appearance imaged by the TIR sensors presents obvious discrimination.
Compared to the straightforward averaging applied in 'w/o DFM', our DFM evaluates the importance of both modalities and provides a better solution, from which the prediction will be more accurate.

\subsection{Ablation Study}
To validate the effectiveness of each component proposed in our method, an ablation study is conducted on two challenging datasets, including VOT-RGBT2019 \cite{VOT2019} and GTOT \cite{GTOT}.
TABLE \ref{res:ablation} reports the quantitative results.
Generally, focusing on the most significant metrics on VOT-RGBT2019 and GTOT, \ie EAO and Success, it is verified that both of our DFM and TIAM modules contribute to the performance improvement.
\begin{table}[ht]
\centering
\caption{\label{res:dfm-network}Exploring the width of our DFM on VOT-RGBT2019 dataset. The top three results are emphasised in \textcolor{red}{red}, \textcolor{blue}{blue} and \textcolor{yellow}{yellow}.}
\begin{tabular}{c|c|ccc}
\toprule
\toprule
Index & Configuration & A ($\uparrow$) & R ($\downarrow$) & EAO ($\uparrow$) \\
\toprule
1 & 1-16-8-1       & 0.6388 & 0.2915 & 0.3980  \\
\midrule
2 & 1-32-16-1      & 0.6400 & 0.2833 & {\color[HTML]{3166FF} 0.4091}  \\
\midrule
3 & 1-64-32-1      & {\color[HTML]{3166FF} 0.6464} & {\color[HTML]{FFC702} 0.2702} & {\color[HTML]{FFC702} 0.4061} \\
\midrule
4 & 1-128-64-1     & 0.6391 & 0.2957 & 0.4057                        \\
\midrule
5 & 1-16-8(4)-1    & {\color[HTML]{FFC702} 0.6446} & 0.2745 & 0.3982 \\
\midrule
6 & 1-32-16(8)-1   & 0.6426 & {\color[HTML]{3166FF} 0.2647} & {\color[HTML]{FE0000} 0.4164} \\
\midrule
7 & 1-64-32(16)-1  & 0.6392 & 0.2989 & 0.4028 \\
\midrule
8 & 1-128-64(32)-1 & {\color[HTML]{FE0000} 0.6485} & {\color[HTML]{FE0000} 0.2545} & 0.4037 \\
\bottomrule
\bottomrule
\end{tabular}
\end{table}

Specifically, on VOT-RGBT2019, DFM brings an improvement of 1.22\% on EAO by adaptively fusing the multi-modal characteristics.
Compared to the baseline tracker, a considerable enhancement (2.84\%) is obtained by our TIAM, which laterally validates the necessity of taking temporal information into account.
Further combining DFM and TIAM, the performance of our method is boosted again, reaching 0.4164 on EAO.
On GTOT, the gains of DFM and TIAM are 1.0\% and 0.7\% separately.
After combining together, our method, TAAT, exceeds the baseline RGBT tracker by 1.8\%.

\begin{figure*}
	\begin{center}
		\includegraphics[width=1\linewidth,height=0.4\linewidth]{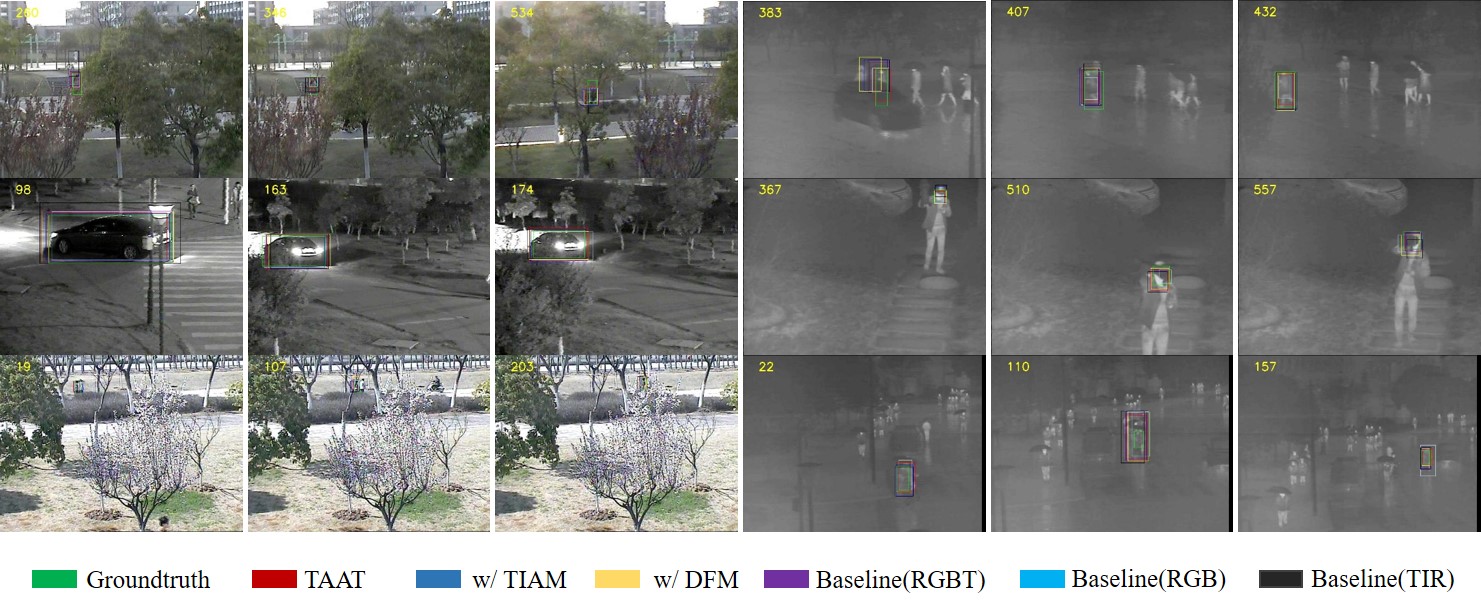} 
	\end{center}
    \caption{Qualitative verification of each component proposed in our method. From top to bottom and left to right, there are some examples from sequence \textit{car37}, \textit{woman89}, \textit{car41}, \textit{face1}, \textit{diamond} and \textit{biketwo}. Due to the small size of objects, this figure is supposed to be viewed after zoomed in.}
    \label{fig:ablation-study}
\end{figure*}

Fig. \ref{fig:ablation-study} illustrates the effectiveness of each component of our method qualitatively.
We show the tracking results both in RGB (\textit{car37}, \textit{car41}, \textit{diamond}) and TIR (\textit{woman89}, \textit{face1}, \textit{biketwo}) modalities.

\section{Conclusion}
In this paper, we propose an adaptive RGBT tracker with temporal information considered (TAAT).
Constraint to the limited available RGBT data, the cross-modal fusion task is achieved at the decision level by an adaptive and lightweight fusion sub-network (DFM).
Furthermore, considering the great significance of temporal clues for video analysis, we extend the construction of the appearance model from only spatial information included to a spatio-temporal manner (TIAM).
Besides, the exhaustive experiments on several benchmarks, \ie VOT-RGBT2019, GTOT and RGBT210, demonstrate the effectiveness and advantages of our approach.

In the future, compared with the short-term construction designed in this method, long-term temporal transformation should be further considered and complemented to the short-term structure, which will contribute to a more robust tracker.


\section*{Acknowledgement}
This work was supported in part by the National Key Research and Development Program of China under Grant 2017YFC1601800, in part by the National Natural Science Foundation of China (Grant NO.62020106012, U1836218, 61672265) and the 111 Project of Ministry of Education of China (Grant No.B12018).

\bibliographystyle{IEEEtran} 
\bibliography{arxiv} 
\end{document}